\documentclass[lettersize,journal]{IEEEtran}
\usepackage{amsmath,amsfonts}
\usepackage{algorithmic}
\usepackage{algorithm}
\usepackage{array}
\usepackage[caption=false,font=normalsize,labelfont=sf,textfont=sf]{subfig}
\usepackage{textcomp}
\usepackage{stfloats}
\usepackage{url}
\usepackage{verbatim}
\usepackage{graphicx}
\usepackage{cite}

\usepackage{multirow}
\usepackage{hyperref}  
\usepackage{cleveref}
\usepackage{pdfpages}  
\usepackage{svg}  
\usepackage{booktabs}
\usepackage[table,xcdraw]{xcolor}  
\usepackage{amssymb}  
\usepackage{pifont}
\usepackage{orcidlink} 

\hypersetup{
    colorlinks=true,        
    linkcolor=blue,         
    citecolor=blue,         
    urlcolor=black           
}

\renewcommand{\cite}[1]{\textcolor{blue}{[\citeonline{#1}]}}  

\newcommand{\revised}[1]{#1}

\begin{document}

\title{SFFNet: Synergistic Feature Fusion Network With Dual-Domain Edge Enhancement for UAV Image Object Detection}

\author{Wenfeng Zhang$^{\orcidlink{0000-0001-7459-2510}}$, Jun Ni$^{\orcidlink{0009-0000-8261-1494}}$, Yue Meng$^{\orcidlink{0009-0001-9952-2348}}$, Xiaodong Pei$^{\orcidlink{0009-0008-1869-3550}}$, Wei Hu$^{\orcidlink{0000-0002-4637-8995}}$, Qibing Qin$^{\orcidlink{0000-0001-7976-318X}}$ and Lei Huang$^{\orcidlink{0000-0003-4087-3677}}$
\thanks{This work was supported in part by Brain-Gain Plan of New Chongqing Foundation under Grant CSTB2024YCJH-KYXM0108, in part by Natural Science Foundation of Chongqing under Grant CSTB2025NSCQ-GPX1037 and CSTB2023NSCQ-MSX0407, in part by National Natural Science Foundation of China under Grant 62302338, in part by Natural Science Foundation Innovation and Development Joint Fund Project of Chongqing under Grant CSTB2023NSCQ-LZX0148, in part by Science and Technology Research Program of Chongqing Municipal Education Commission under Grant KJQN202500532 and KJQN202500511, in part by Chongqing Normal University Foundation under Grant 24XLB018, and in part by Shandong Province Natural Science Foundation under Grant ZR2022QF046 and ZR2025MS1067. \textit{(Wenfeng Zhang and Jun Ni are co-first authors.)} \textit{(Corresponding authors: Wei Hu; Qibing Qin.)}}
\thanks{Wenfeng Zhang, Jun Ni, Wei Hu are with the College of Computer and Information Science, Chongqing Normal University, Chongqing 401333, China (e-mail: itzhangwf@cqnu.edu.cn; 2023210516058@stu.cqnu.edu.cn; wei.workstation@gmail.com).}
\thanks{Yue Meng, Xiaodong Pei are with CETC Yizhihang (Chongqing) Technology Co., Ltd, Chongqing 400031, China (e-mail: yue.neepu@aliyun.com; youngeast@126.com).}
\thanks{Qibing Qin is with the School of Computer Engineering, Weifang University, Weifang 261061, China (e-mail: qinbing@wfu.edu.cn).}
\thanks{Lei Huang is with the Faculty of Information Science and Engineering, Ocean University of China, Qingdao 266005, China (e-mail: huangl@ouc.edu.cn).}
}

\markboth{Journal of \LaTeX\ Class Files,~Vol.~14, No.~8, August~2021}%
{Shell \MakeLowercase{\textit{et al.}}: A Sample Article Using IEEEtran.cls for IEEE Journals}


\maketitle

\begin{abstract}
Object detection in unmanned aerial vehicle (UAV) images remains a highly challenging task, primarily caused by the complexity of background noise and the imbalance of target scales. Traditional methods easily struggle to effectively separate objects from intricate backgrounds and fail to fully leverage the rich multi-scale information contained within images. To address these issues, we have developed a synergistic feature fusion network (SFFNet) with dual-domain edge enhancement specifically tailored for object detection in UAV images. Firstly, the multi-scale dynamic dual-domain coupling (MDDC) module is designed. This component introduces a dual-driven edge extraction architecture that operates in both the frequency and spatial domains, enabling effective decoupling of multi-scale object edges from background noise. Secondly, to further enhance the representation capability of the model's neck in terms of both geometric and semantic information, a synergistic feature pyramid network (SFPN) is proposed. SFPN leverages linear deformable convolutions to adaptively capture irregular object shapes and establishes long-range contextual associations around targets through the designed wide-area perception module (WPM). Moreover, to adapt to the various applications or resource-constrained scenarios, six detectors of different scales (N/S/M/B/L/X) are designed. Experiments on two challenging aerial datasets (VisDrone and UAVDT) demonstrate the outstanding performance of SFFNet-X, achieving 36.8 AP and 20.6 AP, respectively. The lightweight models (N/S) also maintain a balance between detection accuracy and parameter efficiency. The code will be available at \href{https://github.com/CQNU-ZhangLab/SFFNet}{https://github.com/CQNU-ZhangLab/SFFNet}.

\end{abstract}


\section{Introduction}

\begin{figure}[!t]  
\captionsetup{skip=0pt}  
\centering
\includegraphics[width=\columnwidth]{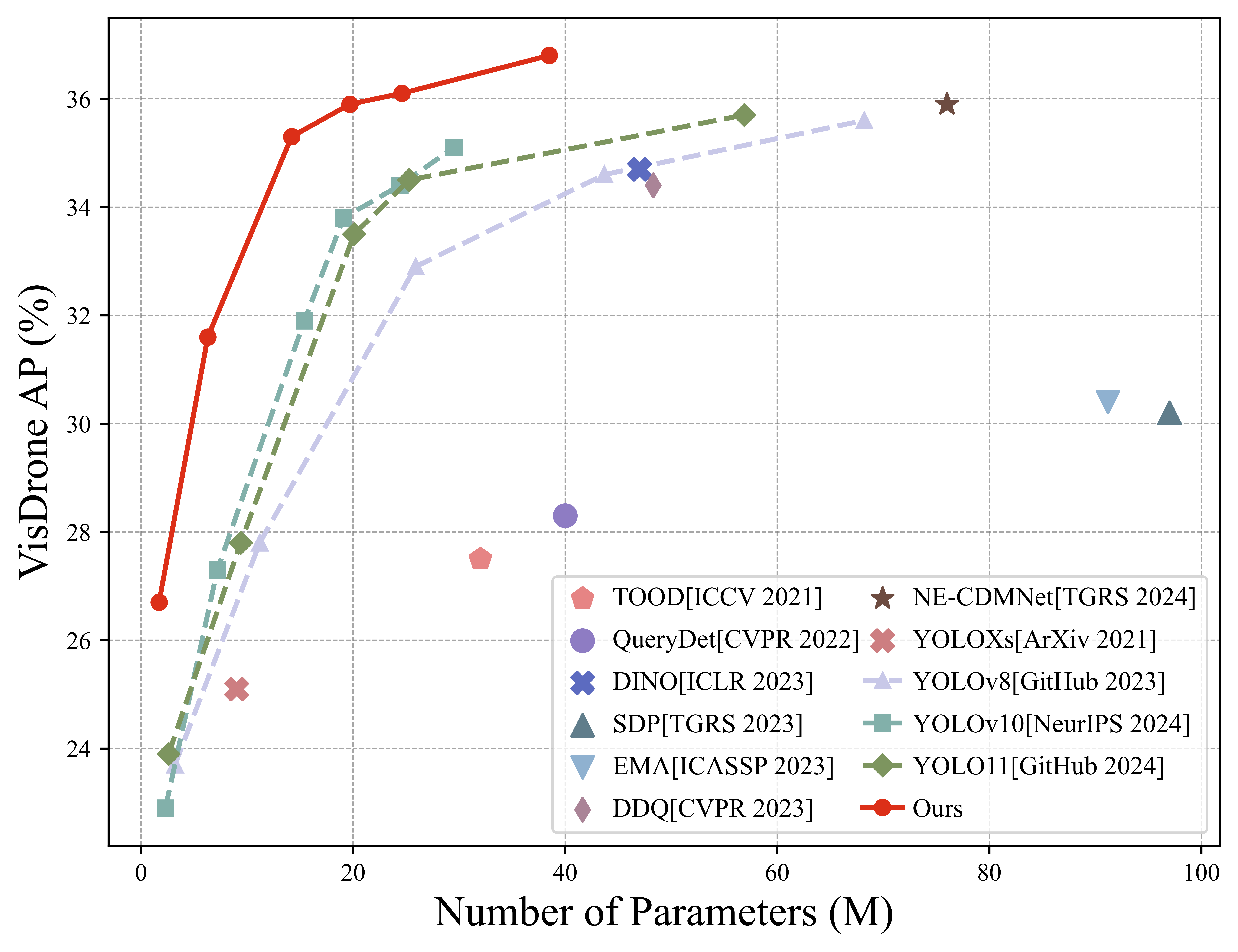}
\caption{\revised{The relationship between the AP value and the number of parameters for different object detection algorithms on the VisDrone dataset. The points near the upper left corner of the figure indicate that the model can achieve higher accuracy while maintaining a lower number of parameters. Our algorithm (marked “Ours”) outperforms others by achieving a higher AP value with fewer parameters.}}
\label{fig_1}
\end{figure}

\IEEEPARstart{I}{n} aerial image object detection tasks, unmanned aerial vehicle (UAV) have become an important platform for efficient monitoring due to their high flexibility and wide field of view. Their applications in many fields such as urban monitoring \cite{ref91}, \cite{ref90}, \cite{ref89}, \cite{ref93}, \cite{ref92}, military reconnaissance \cite{ref21}, environmental protection \cite{ref18} and agricultural management \cite{ref94}, are becoming more and more widespread. UAV aerial images, typically captured from elevated altitudes and diverse angles, are prone to dynamic backgrounds and environmental noise. These images exhibit a wide range of object scales, with a predominance of smaller objects, which exacerbates the presence of background interference. The scale imbalance and dominance of noise challenge the generalization ability of object detection models, while the limited information for small objects further complicates feature learning.

Multi-scale feature fusion methods alleviate the scale imbalance problem by combining feature information from different scales. Early architectures, such as the feature pyramid network (FPN) \cite{ref3}, employed end-to-end multi-scale feature construction. However, they overlooked the synergistic relationships among different feature layers. The bi-directional feature pyramid network (BiFPN) \cite{ref1} enhances the correlation between features by introducing a bi-directional weighted fusion mechanism, whereas this mechanism may also lead to potential information loss.

In order to enable the model to better capture fine-grained features, several methods, such as EFPN \cite{ref42}, Spectral-Spatial FPN \cite{ref22}, and CascadeDumpNet \cite{ref25}, have introduced additional detection heads at the P2 layer, and even at the P1 layer, of the neck feature fusion network. While these approaches improve detection accuracy by directly utilizing low-level detailed information, they inevitably impose higher demands on computational resources.

Therefore, in order to address the limitations of existing methods, we improve model performance in the following two aspects. (1) In response to the dual challenges of insufficient multi-scale edge feature representation and limited background noise suppression in current backbone networks, we propose a multi-scale dynamic dual-domain coupling (MDDC) module. This module is integrated into the feature extraction stage of the backbone network, overcoming the limitations of traditional single-domain convolution by constructing a frequency-domain and spatial-domain dual-drive architecture. MDDC utilizes frequency domain analysis to achieve the spectral space decoupling of multi-scale target edge features and background noise. By selectively suppressing specific frequency bands, it eliminates interference components, thereby enabling precise edge information extraction and enhancing noise immunity. (2) To address the challenges of inadequate representation of small target features and the excessive computational burden imposed by redundant detection heads within existing feature pyramids, we propose the synergistic feature pyramid network (SFPN). The SFPN leverages linear deformable convolution to establish a low-level feature correlation pathway, in lieu of incorporating detection heads at the lower layers. This innovative design avoids the increased computational burden from adding extra detection heads, while effectively utilizing the dynamic kernel deformation mechanism to precisely localize the geometric structure of small targets. Furthermore, the wide-area perception module (WPM) is deployed at the P3 level, utilizing parallel heterogeneous convolutional kernels to capture contextual information across receptive fields, thereby enhancing the target's high-level semantic representation capability.

In Fig.~\ref{fig_1}, we provide a comparison of our method with several state-of-the-art object detectors. An analysis of the trends in the curves unequivocally reveals that our detector attains superior detection accuracy, while utilizing a comparable or reduced number of parameters. The contributions of this article can be summarized as follows:

1) A novel multi-scale dynamic dual-domain coupling (MDDC) module is proposed to address feature confusion induced by complex background noise in UAV images, as well as detection bias arising from object scale imbalance. MDDC integrates frequency-domain edge enhancement with spatial-domain feature refinement, effectively ensuring the precise separation of object edges from intricate backgrounds through a multi-scale dynamic coupling mechanism.

2) An efficient synergistic feature pyramid network (SFPN) is proposed. We thoroughly explore the potential of feature pyramids within the neck to represent both geometric and long-range contextual information, thereby enabling synergistic multi-scale object modeling with SFPN.

3) To adapt to the various applications or resource-constrained scenarios, six models of different scales are designed (N/S/M/B/L/X). Their outstanding performance is validated through both quantitative and qualitative experiments on the VisDrone and UAVDT datasets, demonstrating the adaptability and scene coverage capabilities of the proposed method.

This article is organized in the following manner. Initially, in Section \ref{sec:2}, a brief review of related work is provided, exploring various established methodologies in the field. Subsequently, in Section \ref{sec:3}, the proposed method is elaborated, detailing the underlying principles. \revised{Following this, Section \ref{sec:4} introduces the datasets used and details the experimental setup, followed by an evaluation of the proposed approach.} Finally, in Section \ref{sec:5}, the article concludes by summarizing the key findings and discussing potential future research directions.

\section{Related Work}
\label{sec:2}

\subsection{Aerial Image Object Detection}
Aerial image object detection encounters several challenges, including complex backgrounds, varying target scales, and perspective distortions. To mitigate these issues, some researchers have proposed combining coarse and fine detection strategies. For instance, Yang et al. \cite{ref39} proposed ClusDet, which generates candidate boxes for clustering and applies scale normalization and fine detection. Li et al. \cite{ref40} introduced DMNet, using density maps for crowd counting to guide image cropping and optimize detection regions. Huang et al. \cite{ref41} developed UFPMP-Det, a multi-proxy detection network, where a coarse detector generates subregions that are merged to minimize background noise. These methods depend on the integration of coarse and fine detection, which can lead to elevated computational demands and reduced processing efficiency. Furthermore, the coarse detector has the potential to generate false positives in densely populated areas, undermining the performance of the fine detector.

Some researchers have also utilized contextual information from the surrounding scene, leading to more optimal feature representations and subsequently enhancing detection accuracy. For instance, SCDNet \cite{ref46} decouples scene contextual information using a dedicated scene classification subnetwork (SCN), thereby enhancing the modeling of the relationships between dense small objects and their surroundings. Additionally, HRFG \cite{ref50} uses high-resolution features and a background degradation strategy to improve small object detection through image super-resolution. However, aerial image detection methods relying on contextual enhancement can suffer from performance instability in complex scenes, while super-resolution techniques often come with high computational burden and limited generalization, especially in low-resolution images.

\subsection{Frequency Domain Learning}
Frequency domain techniques have been progressively developed in the field of object detection in satellite remote sensing imagery. To further improve feature extraction and detection performance, methods that integrate frequency domain information into CNNs have been introduced. For instance, Weng et al. \cite{ref30} proposed a selective frequency-domain interaction (SFI) network that integrates both frequency and time domains, addressing frequency domain interaction challenges. In a similar vein, Zheng et al. \cite{ref53} developed an instance-oriented spatial-frequency-domain feature fusion detector (SFFD) to address the lack of directional information in CNNs. Moreover, Zheng et al. \cite{ref51} proposed a frequency-domain feature extraction (FFE) network designed to capture frequency features across spatial positions. An orientation-enhanced self-attention layer (OES-Layer) was incorporated to learn the directional information of targets, which enhances the spatial response of target instances. Compared to remote sensing images, Wang et al. \cite{ref56} proposed an innovative frequency-domain decoupling method to enhance the generalization ability of UAV object detection. This method utilizes learnable filters to extract domain-invariant and domain-specific spectra and designs image-level and instance-level contrast losses to guide training, achieving improved frequency domain decoupling.

Although frequency domain techniques in both fields utilize frequency domain feature refinement, they fail to fully exploit its potential, resulting in inaccurate feature extraction that impedes the capture of detailed features. In contrast, MDDC enables the model to harness both frequency and spatial domains for dual-driven edge extraction, leading to the decoupling of multi-scale target edges from background noise and subsequently improving edge information extraction and noise resistance.

\begin{figure*}[!t]   
\captionsetup{skip=0pt}  
\centering
\includegraphics[width=1\textwidth]{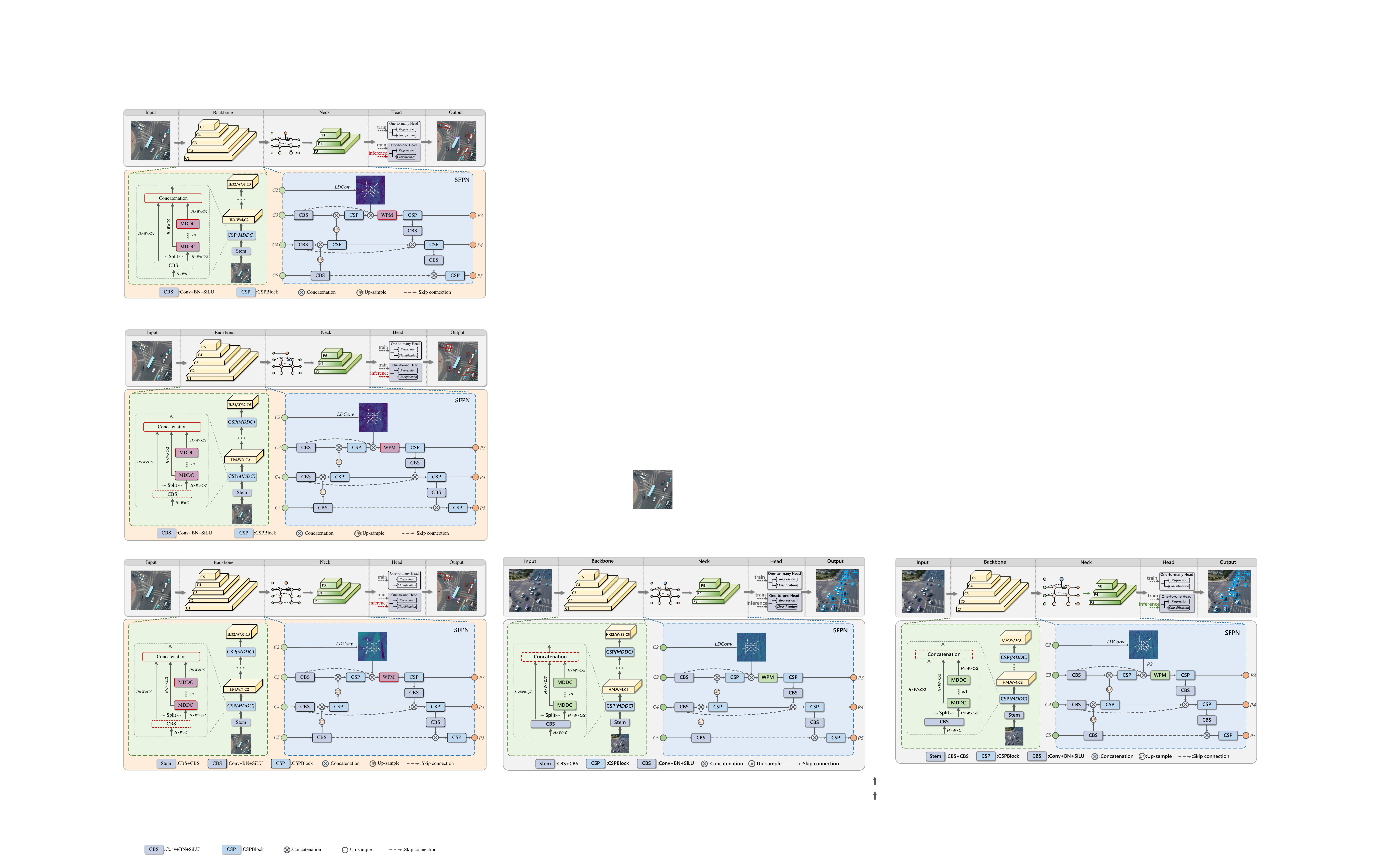}  
\caption{\revised{Overview of the SFFNet framework. The framework integrates a backbone network with MDDC for efficient multi-scale dual-domain feature extraction, and realizes collaborative feature fusion in the neck through SFPN. This design ensures the accurate positioning and detection of small objects, effectively addressing the challenges in complex environments.}}
\label{fig_4}
\end{figure*}

\subsection{Feature Enhancement and Multi-Scale Fusion}
Convolutional Neural Networks (CNNs) generate feature maps with varying spatial resolutions as a result of deep architectures. The shallow layers are responsible for capturing fine-grained details and localization cues, whereas the deeper layers extract rich semantic information. However, due to subsampling operations, such as stride convolutions and pooling layers, the responses corresponding to small objects tend to be attenuated in deeper features. Therefore, it is imperative to develop effective strategies for feature enhancement and multi-scale fusion to address this challenge.

The feature pyramid network (FPN) \cite{ref3} aggregates multi-scale features through a top-down path, improving detection performance for objects of varying sizes. PAFPN \cite{ref4} extends FPN by adding a bottom-up path to better capture and integrate features at different scales. EfficientDet \cite{ref1} introduces BiFPN, which enhances feature representation with residual connections and weighted fusion for richer spatial information. AugFPN \cite{ref27} optimizes multi-scale feature use through consistency supervision, residual enhancement, and soft RoI selection. AFPN \cite{ref86} improves traditional pyramid networks by enabling direct interaction between non-adjacent levels. DN-FPN \cite{ref2} uses contrastive learning to reduce noise and preserve geometric and semantic details. SAFPN \cite{ref35} adds a spatially aware alignment fusion module to mitigate spatial misalignment, enhancing fine-grained feature extraction. While these methods optimize the feature representation problem during the feature fusion process, the spatial details of low-level features are not fully utilized, and the modeling of long-range contextual information is neglected when constructing multi-scale feature couplings.

The SFPN we have developed employs linear deformable convolution to integrate low-level information from the backbone, eliminating the need for extra detection heads. This design captures detailed spatial information while controlling parameter count. The WPM at the P3 layer enhances contextual modeling in high-resolution feature maps, and the concatenation fusion strategy reduces learning bias from insufficient small object feedback.

\section{Proposed Method}
\label{sec:3}
In this section, we will first introduce an overview of the SFFNet framework, and then introduce our proposed MDDC, SFPN, and WPM in turn.

\begin{figure*}[!t]   
\captionsetup{skip=0pt}  
\centering
\includegraphics[width=1\textwidth]{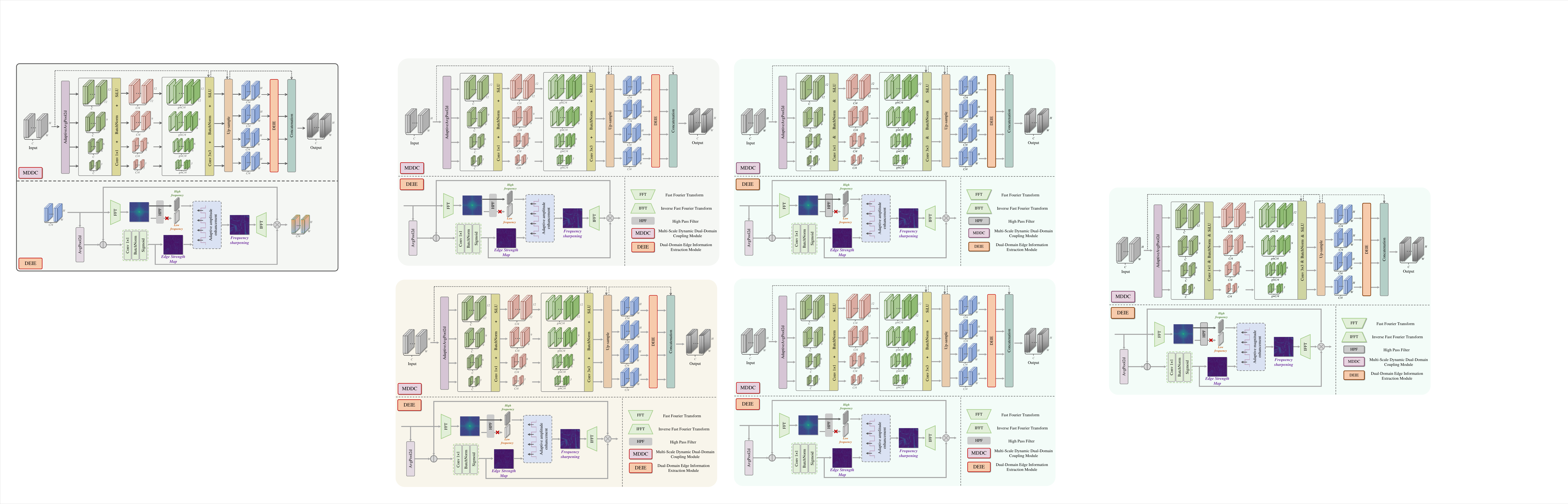}  
\caption{The detailed structure of the MDDC module. The MDDC module initially performs multi-scale decomposition on the input feature map to construct the base representation. Subsequently, through DEIE, it achieves complementary feature fusion in both the spatial and frequency domains. In the frequency domain branch of DEIE, features undergo spectral decomposition, removing low-frequency noise disturbances while enhancing high-frequency feature representations.}
\label{fig_5}
\end{figure*}

\subsection{Framework Overview}
As shown in Fig.~\ref{fig_4}, SFFNet first extracts multi-scale feature maps from the original input image through a backbone network mainly composed of a series of convolutions and MDDC modules. Next, the SFPN is used to perform further feature processing on the feature maps at different scales. \revised{In particular, the C2 layer features are first processed by a linear deformable convolution and then forwarded to the P3 layer, where they are further enhanced by the WPM module.} Finally, the semantic category and spatial coordinates of the target are decoded in parallel by two detection heads. Among them, the one-to-many detection head is only effective during training and is turned off during inference.

\subsection{Multi-Scale Dynamic Dual-Domain Coupling}
Compared to traditional spatial-domain convolution methods, frequency-domain techniques can more precisely distinguish high-frequency features of target edges from low-frequency background noise through spectral decomposition mechanisms. Leveraging the advantages of frequency-domain technology in feature decoupling, and combining it with spatial-domain methods, we have designed a multi-scale dynamic dual-domain coupling (MDDC) module. MDDC enhances the model's capability to represent fine details of small object edges by employing operations including multi-scale feature generation and dual-domain edge information extraction (DEIE). Its structure is depicted in Fig.~\ref{fig_5}.

\subsubsection{Multi-Scale Feature Generation}
The MDDC first takes the feature map \( X \in \mathbb{R}^{C \times H \times W} \) as input, and then generates feature maps with different sizes by applying adaptive average pooling operations. This process can be formulated as
\begin{equation}
X_s = \mathrm{AAP}_s(X),
\end{equation}
where \( s \in \{3, 6, 9, 12\} \) represents the size of the feature map after pooling.

Then, the feature maps of each scale are sequentially processed through two sets of convolution operations (\(\mathrm{Conv}1\times1\) and \(\mathrm{GroupConv}3\times3\)). In the first part, the \(1\times1\) convolution reduces the number of channels in the feature map to a quarter of the original, extracting more concise features. Specifically, the \(3\times3\) grouped convolution adopts a channel decoupling strategy, performing independent spatial convolution operations on each channel, effectively maintaining the channel-specific characteristics of the feature representation. This process can be described as
\begin{equation}
    X_s^{\mathrm{conv}1}=\mathrm{SiLU}\left(\mathrm{BN}(\mathrm{Conv}1\times1(X_s))\right),
\end{equation}
\begin{equation}
    X_s^{\mathrm{conv}3}=\mathrm{SiLU}\left(\mathrm{BN}(\mathrm{GroupConv}3\times3(X_s^{\mathrm{conv}1}))\right).
\end{equation}

Through convolution, the extracted feature map \(X_s^{\mathrm{conv}3}\) is up-sampled to restore the resolution of the original input feature map \({X}\). The aligned multi-scale feature map \(X_{\mathrm{up}}\) is then fed into the DEIE module to extract edge information from the features.

\subsubsection{Dual-Domain Edge Information Extraction (DEIE)}
Forward propagation of the DEIE module first smooths the input feature map \(X_{\mathrm{up}}\in\mathbb{R}^{\frac{C}{4}\times H\times W}\) by passing it through a \(3\times3\) average pooling layer to calculate the low-frequency component. Then, the low-frequency feature component is subtracted from the original input feature map \(X_{\mathrm{up}}\) to extract the high-frequency component \(X_{\mathrm{high}}\). This is then further processed by a set of \(1\times1\) convolution layers. 
\begin{equation}
    X_{\mathrm{high}}(i,j)=X_{\mathrm{up}}(i,j)-\frac{1}{k^{2}}\sum_{m=-1}^{1}\sum_{n=-1}^{1}X_{\mathrm{up}}(i+m,j+n),
\end{equation}
where \((i,j)\) represents the current pixel location. \({m}\) and \({n}\) are the offsets within the window. The size of the pooling kernel is \(k=3\). 

On the other hand, in order to extract specific high-frequency components, the original feature map \(X_{\mathrm{up}}\) will be converted from the spatial domain to the frequency domain by a two-dimensional discrete fourier transform. This transformation can be described as
\begin{equation}
    \begin{aligned}
    X_{\mathrm{up}}(u,v)&=\mathcal{F}\{X_{\mathrm{up}}(i,j)\}\\&=\sum_{i=0}^{M-1}\sum_{j=0}^{N-1}X_{\mathrm{up}}(i,j)e^{-\mathrm{i}\cdot2\pi\left(\frac{ui}{M}+\frac{vj}{N}\right)},
    \end{aligned}
\end{equation}
where \(X_{\mathrm{up}}({u,v})\) is the frequency domain representation after the fourier transform. \(X_{\mathrm{up}}({i,j})\) represents the pixel value of the input 2D spatial domain image, and \({i}\) and \({j}\) represent the row and column coordinates of the image in the spatial domain, respectively. \(e^{-\mathrm{i}\cdot2\pi\left(\frac{ui}{M}+\frac{vj}{N}\right)}\) represents the kernel function of the fourier transform, the exponential part \(-\mathrm{i}\cdot2\pi\left(\frac{ui}{M}+\frac{vj}{N}\right)\) represents the phase rotation, \({M}\) and \({N}\) represent the width and height in the spatial domain, and \(\mathrm{i}\) is the imaginary unit. \(\frac{ui}{M}\) and \(\frac{vj}{N}\) represent the spatial variation of the image in the horizontal and vertical directions at frequencies \({u}\) and \({v}\), respectively, which are used to modulate the pixel values in the \({i}\) and \({j}\) directions.

In the frequency domain, we apply a high-pass filter to retain high-frequency information and suppress low-frequency components. Specifically, frequencies below the threshold are reset to zero while retaining the original phase information to avoid distorting the overall structure of the image. This process can be described as
\begin{equation}
    \begin{aligned}
    \mathcal{M}\left(X_{\mathrm{up}}(u,v)\right) &= \begin{vmatrix}X_{\mathrm{up}}(u,v)\end{vmatrix} \\
    &= \sqrt{\mathrm{Re}(X_{\mathrm{up}}(u,v))^{2} + \mathrm{Im}(X_{\mathrm{up}}(u,v))^{2}} \raisebox{-0.5ex}{,}
    \end{aligned}
\end{equation}
\begin{equation}
    \mathcal{P}(X_{\mathrm{up}}(u,v))=\arctan\left(\frac{\mathrm{Im}(X_{\mathrm{up}}(u,v))}{\mathrm{Re}(X_{\mathrm{up}}(u,v))}\right),
\end{equation}
\begin{equation}
    \mathrm{M}_{\mathrm{hpf}}(u,v)=\begin{cases}\mathcal{M}\left(X_{\mathrm{up}}(u,v)\right),\!\!\!\!\quad\mathrm{if}\!\!\!\!\quad\mathcal{M}\left(X_{\mathrm{up}}(u,v)\right)\geq\alpha\\\quad0,\!\!\!\!\quad\mathrm{otherwise}&\end{cases}\!\!\!\!\!\!,
\end{equation}
\begin{equation}
    X_{\mathrm{hpf}}(u,v)=\mathrm{M}_{\mathrm{hpf}}(u,v)\cdot e^{\mathrm{i}\cdot \mathcal{P}(X_{\mathrm{up}}(u,v))},
\end{equation}
where \(\mathcal{M}\left(X_{\mathrm{up}}(u,v)\right)\) and \(\mathcal{P}\left(X_{\mathrm{up}}(u,v)\right)\) represent the magnitude and phase of the frequency domain image \(X_{\mathrm{up}}(u,v)\) at position \((u,v)\), respectively. \(X_{\mathrm{hpf}}(u,v)\) is the frequency domain reconstructed image after high-pass filtering. In this study, the magnitude threshold \(\alpha\) is set to 0.1.

When processing the frequency-domain image \(X_{\mathrm{hpf}}(u,v)\) after high-pass filtering, it is necessary to further adaptively enhance the magnitude of specific frequency components based on the edge strength of the image content. We use the edge strength map of the high-frequency components \(X_{\mathrm{high}}\) as a guide to enhance the magnitude, while ensuring that the structural features of the reconstructed image are not distorted. The edge strength map is obtained by calculating the local contrast of the high-frequency components \(X_{\mathrm{high}}\), depicting the intensity variations in local regions of the image. This process can be described as
\begin{equation}
    \mathrm{S}(i,j)=|X_{\mathrm{high}}(i,j)-\frac{1}{r^{2}}\sum_{p=-1}^{1}\sum_{q=-1}^{1}X_{\mathrm{high}}(i+p,j+q)|,
\end{equation}
\begin{equation}
    \mathrm{ME}(u,v)=\begin{vmatrix}X_\mathrm{hpf}(u,v)\end{vmatrix}\cdot\begin{pmatrix}1+\beta\cdot\mathrm{S}(i,j)\end{pmatrix},
\end{equation}
\begin{equation}
    X_{\mathrm{e}}(u,v)=\mathrm{ME}(u,v)\cdot e^{\mathrm{i}\cdot \mathcal{P}(X_{\mathrm{hpf}}(u,v))},
\end{equation}
where \(\mathrm{S}(i,j)\) represents the pixel value at position \((i,j)\) in the edge strength image. \({r}\) represents the domain size factor 3. \(\mathrm{ME}(u,v)\) represents the magnitude after edge strength image guided enhancement. \(X_{\mathrm{e}}(u,v)\) represents the reconstruction in the frequency domain after magnitude enhancement with phase invariance. In this article, the intensity factor \(\mathrm{\beta}\) is set to 1.5.

Afterwards, perform frequency sharpening on the adaptively enhanced frequency-domain image \( X_{e}(u,v) \). This process can be formulated as
\begin{equation}
    X_{\mathrm{fs}}(u,v)=|X_{e}(u,v)|\cdot\gamma\cdot e^{\mathrm{i}\cdot \mathcal{P}(X_{e}(u,v))},
\end{equation}
where \( X_{\mathrm{fs}}(u,v) \) is the sharpened frequency-domain image, and \( \gamma \) is the sharpening factor set to 1.2. 

Subsequently, the frequency-sharpened part \( X_{\mathrm{fs}}(u,v) \) is transformed back to the spatial domain using a 2D discrete inverse fourier transform to obtain the enhanced component \( X_{\mathrm{fs}}(i,j) \). This transformation can be described as
\begin{equation}
    \begin{aligned}
    X_{\mathrm{fs}}(i,j) &= \mathcal{F}^{-1}\{X_{\mathrm{fs}}(u,v)\} \\
    &= \frac{1}{MN}\sum_{u=0}^{M-1}\sum_{v=0}^{N-1}X_{\mathrm{fs}}(u,v)e^{\mathrm{i}\cdot2\pi\left(\frac{ui}{M}+\frac{vj}{N}\right)}, 
    \end{aligned}
\end{equation}

Finally, the original input image \(X_{\mathrm{up}}\), the high-frequency component \(X_{\mathrm{high}}\) and the sharpening component \(X_{\mathrm{fs}}\) are spliced and output through three learnable adaptive weight branches.

In addition, MDDC retains a branch without bi-domain edge enhancement to ensure the integrity of the spatial domain discriminative features and reduce possible misunderstandings or biases caused by a single source of information. This branch only contains a set of convolution operations (\(\mathrm{GroupConv}3\times3\)). Subsequently, all feature maps will be spliced and integrated, and finally feature fusion will be completed through a 1 × 1 convolution layer to generate the final output feature map.

\subsection{Synergistic Feature Pyramid Network}

The detailed structure of SFPN is shown in Fig.~\ref{fig_4}. SFPN achieves efficient feature representation and enhancement by progressively fusing multi-scale features layer by layer. It begins with high-level features, incorporates lower-level ones, and performs bottom-up fusion. Concatenation is used for fusion to retain complete information without altering feature ratios. At the P2 level, linear deformable convolution enhances fine detail extraction, emphasizing small object information. At P3, the wide-area perception module (WPM) captures both detailed object features and contextual information.

\subsubsection{Linear Deformable Convolution (LDConv)}
From the perspective of UAVs, the shape of objects is often diverse, especially in object detection tasks, where the shape of an object may undergo significant changes due to factors such as different viewpoints, distances, rotations, and scale variations. Traditional convolution operations typically rely on fixed kernels, which restrict their capacity to capture variations in shape. As an object's shape undergoes transformation, these convolutions often face difficulties in adaptation.

LDConv \cite{ref5} dynamically adjusts the sampling locations of the kernel, enabling the adaptive capture of object shape changes while ensuring a linear increase in parameter count. \revised{Based on these advantages and the trend of feature detail distribution in the feature pyramid, we apply LDConv to the C2 feature map from the backbone, which serves as the lowest-level input in the top-down path aggregation of the SFPN.} This design allows for the extraction of more information from small objects in low-level features, thereby enhancing the spatial expressiveness and detail-capturing ability of the features.

\begin{figure}[!t]   
\captionsetup{skip=0pt}  
\centering
\includegraphics[width=\columnwidth]{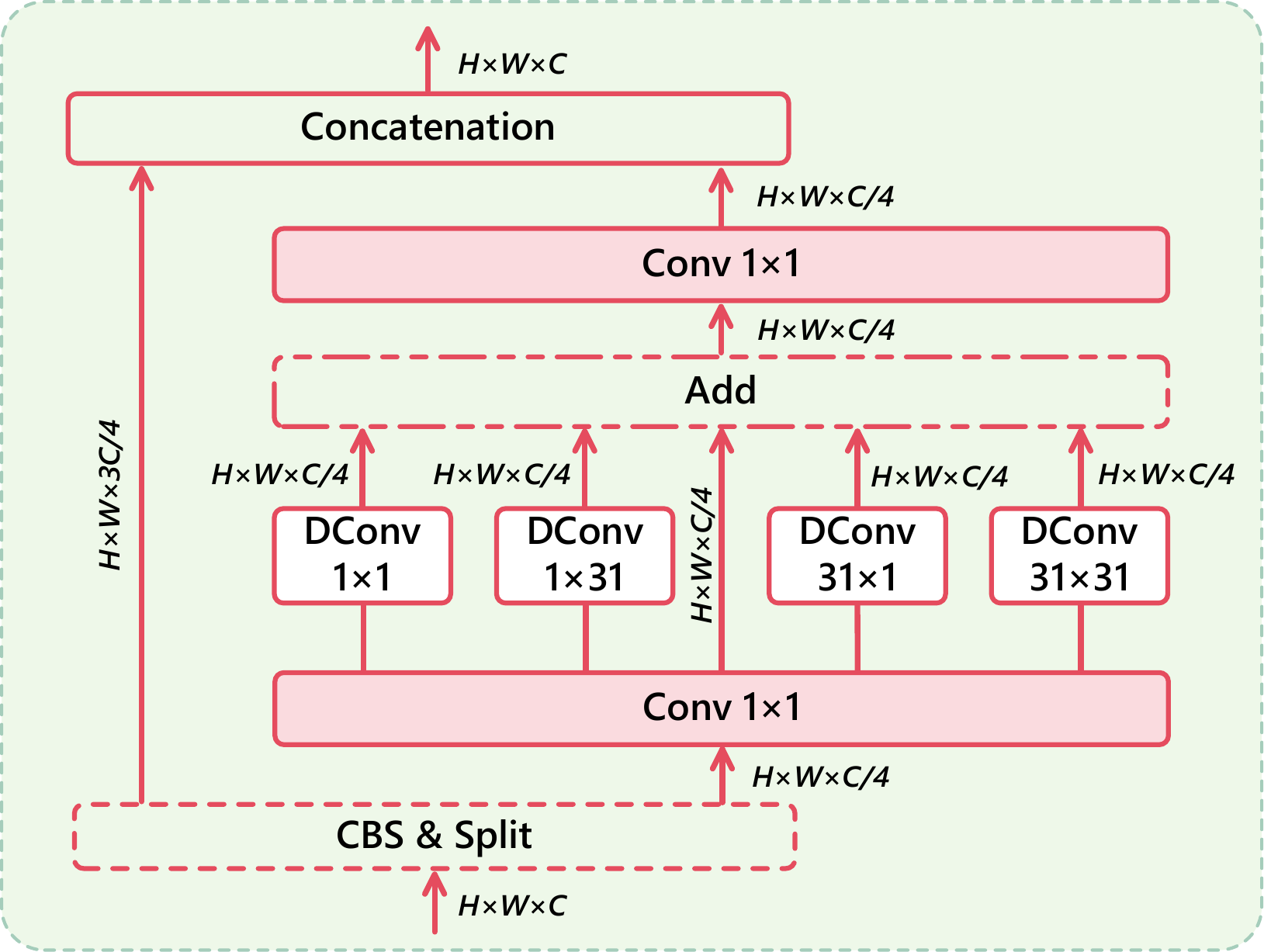}  
\caption{The detailed structure of the WPM. WPM employs a parallel structure consisting of a large kernel convolution, a small kernel convolution, and two strip convolutions.}
\label{fig_6}
\end{figure}

\subsubsection{Wide-Area Perception Module (WPM)}
Accurate detection of objects requires extensive contextual information, and the contextual requirements of different objects are also different. Traditional small-scale convolutional kernels are limited by local receptive fields and have difficulty modeling long-range spatial dependencies. In particular, when dealing with targets with a wide aspect ratio or direction sensitivity, global contextual information is easily lost \cite{ref60}. \revised{Remote sensing and aerial imagery often contain elongated structured targets that exhibit strong directional characteristics. To effectively capture such anisotropic features, orthogonal strip convolutions (\(1\times K\) and \(K\times1\)) offer a more parameter-efficient alternative to dense square kernels while preserving long-range dependencies along specific axes \cite{ref61}.} Large strip convolutions are a good feature representation learner for remote sensing object detection. Using a few large convolutional kernels (Scaling Up Kernels to \(31\times31\)) instead of a bunch of small kernels may be a more powerful paradigm \cite{ref61}, \cite{ref62}. Based on this, we designed a WPM at the P3 layer of the feature pyramid, which simultaneously integrates the complementary advantages of large-kernel convolutions and strip convolutions, fully utilizing the unique long-range context information presented in aerial scenes. \revised{Structurally, WPM employs a parallel composition of large-kernel convolutions (\(31\times31\)) to aggregate isotropic global context, orthogonal strip convolutions (\(1\times31\) and \(31\times1\)) to capture anisotropic and direction-sensitive features, and 1×1 convolutions together with an identity path to stabilize feature learning and preserve local intensity anchors. Such a complementary design jointly models wide-range geometric and directional dependencies, allowing WPM to adaptively meet the heterogeneous contextual requirements of various object categories while maintaining computational efficiency.} The design details of the WPM are shown in Fig.~\ref{fig_6}. Considering this practical factor, the feature maps near the lower layers of the feature pyramid have larger sizes, which can lead to an increase in computational load. To reduce the computational burden, we only feed a quarter of the feature channels into a set of parallel depth convolutions for processing.

Specifically, the WPM first processes the input tensor \(X\in\mathbb{R}^{C\times H\times W}\) through a CBS component, and then divides it into two parts, with sizes \(X_1\in\mathbb{R}^{\frac{C}{4}\times H\times W}\) and \(X_2\in\mathbb{R}^{\frac{3C}{4}\times H\times W}\), respectively. Next, \(X_1\in\mathbb{R}^{\frac{C}{4}\times H\times W}\) is processed through a \(1\times1\) convolutional layer. The feature maps obtained after the convolution operation are then fed into a set of parallel depth convolutions, including \(1\times1\), \(1\times31\), \(31\times1\), and \(31\times31\) convolutional layers. In this process, smaller convolutional kernels refine image information by capturing local features, while large-kernel strip convolutions efficiently capture direction-sensitive features through asymmetric structures. Large-kernel convolutions enhance spatial context modeling with a wide field of view. \revised{Strip convolutions (\(1\times31\) and \(31\times1\)) serve as a separable approximation to large square kernels: they retain long-range dependency along a single axis while reducing background interference from irrelevant directions \cite{ref100}.}

Additionally, to prevent feature degradation, we deliberately retain an identity branch, ensuring that the network does not compromise the integrity of the original features while learning incremental features. The processed tensor will be concatenated along the channel dimension. The concatenated tensor will then be fused through a \(1\times1\) convolution to represent the relationships between the channels. Finally, this portion of the tensor will be further merged with the remaining three-quarters of the channel tensors to obtain the final output.

\begin{table*}[!t]
\centering  
\caption{\revised{Comparison of Our Model with Several YOLO Detectors on the VisDrone Validation Set. The Best Result for Each Scale Is Marked in Bold, and $^{\dagger}$ Denotes a Reimplementation of the Results under the Same Experimental Setting}
\label{tab:table1}} 
\renewcommand{\arraystretch}{1.25}  
\resizebox{0.9\textwidth}{!}{  
\begin{tabular}{l||c c c c c c|c c c c c c|c c}  
\toprule
Model & \(\mathrm{AP}\) & \(\mathrm{AP}_{\mathrm{50}}\) & \(\mathrm{AP}_{\mathrm{75}}\) & \(\mathrm{AP}_{\mathrm{s}}\) 
& \(\mathrm{AP}_{\mathrm{m}}\) & \(\mathrm{AP}_{\mathrm{l}}\) & \(\mathrm{AR}_{\mathrm{1}}\) & \(\mathrm{AR}_{\mathrm{10}}\) & \(\mathrm{AR}_{\mathrm{100}}\) & \(\mathrm{AR}_{\mathrm{s}}\) & \(\mathrm{AR}_{\mathrm{m}}\) & \(\mathrm{AR}_{\mathrm{l}}\) & Param(M) & FLOPs(G) \\
\midrule
YOLOv8-N$^{\dagger}$ \cite{ref9} & 23.7 & 40.2 & 24.0 & 16.8 & 32.0 & 34.5 & 9.9 & 28.9 & 38.8 & 31.5 & 49.3 & 48.9 & 3.2 & 8.7 \\
YOLOv10-N$^{\dagger}$ \cite{ref16} & 22.9 & 39.9 & 22.8 & 17.1 & 30.6 & 26.0 & 9.3 & 28.1 & 38.7 & 32.0 & 48.1 & 42.2 & 2.3 & 6.7 \\
YOLO11-N$^{\dagger}$ \cite{ref17} & 23.9 & 40.6 & 23.9 & 16.9 & 32.5 & 33.3 & 9.6 & 29.0 & 38.9 & 31.1 & 50.1 & 48.9 & 2.6 & 6.5 \\
\rowcolor{gray!15}  
\textbf{SFFNet-N (Ours)} & \textbf{26.7} & \textbf{45.4} & \textbf{26.8} & \textbf{19.8} & \textbf{35.3} & \textbf{35.0} & \textbf{10.4} & \textbf{31.5} & \textbf{42.6} & \textbf{35.6} & \textbf{52.8} & \textbf{52.1} & 1.7 & 7.2 \\
\midrule
\midrule
YOLOv8-S$^{\dagger}$ \cite{ref9} & 27.8 & 46.6 & 28.2 & 20.6 & 37.0 & \textbf{35.8} & 11.1 & 32.2 & 42.6 & 35.4 & 53.1 & 48.2 & 11.2 & 28.6 \\
YOLOv10-S$^{\dagger}$ \cite{ref16} & 27.2 & 46.2 & 27.5 & 20.8 & 35.6 & 31.9 & 10.7 & 31.9 & 43.2 & 36.7 & 52.5 & 46.5 & 7.2 & 21.6 \\
YOLO11-S$^{\dagger}$ \cite{ref17} & 27.8 & 46.5 & 28.1 & 20.5 & 37.3 & 34.9 & 11.0 & 32.4 & 42.8 & 35.5 & 53.9 & 50.6 & 9.4 & 21.5 \\
\rowcolor{gray!15}  
\textbf{SFFNet-S (Ours)} & \textbf{31.6} & \textbf{52.8} & \textbf{32.3} & \textbf{25.3} & \textbf{40.8} & 35.7 & \textbf{12.1} & \textbf{35.7} & \textbf{47.4} & \textbf{40.9} & \textbf{57.3} & \textbf{56.7} & 6.3 & 24.0 \\
\midrule
\midrule
YOLOv8-M$^{\dagger}$ \cite{ref9} & 32.9 & 53.3 & 33.9 & 25.0 & 43.6 & 43.4 & 12.8 & 36.5 & 47.3 & 39.9 & 58.6 & 60.7 & 25.9 & 78.9 \\
YOLOv10-M$^{\dagger}$ \cite{ref16} & 31.9 & 52.6 & 32.8 & 24.7 & 42.0 & 40.7 & 12.3 & 35.8 & 47.4 & 40.3 & 57.9 & 54.7 & 15.4 & 59.1 \\
YOLO11-M$^{\dagger}$ \cite{ref17} & 33.5 & 54.4 & 34.6 & 25.8 & 44.1 & \textbf{47.5} & 12.9 & 37.1 & 48.2 & 40.9 & 59.5 & \textbf{62.6} & 20.1 & 68.0 \\
\rowcolor{gray!15}  
\textbf{SFFNet-M (Ours)} & \textbf{35.3} & \textbf{57.5} & \textbf{36.6} & \textbf{28.0} & \textbf{46.0} & 42.9 & \textbf{13.2} & \textbf{38.5} & \textbf{50.4} & \textbf{43.7} & \textbf{60.6} & 57.4 & 14.2 & 63.2 \\
\midrule
\midrule
YOLOv10-B$^{\dagger}$ \cite{ref16} & 33.8 & 55.4 & 34.9 & 26.3 & 44.2 & 38.5 & 12.7 & 37.3 & 49.1 & 42.2 & 59.9 & 57.3 & 19.1 & 92.0 \\
\rowcolor{gray!15}  
\textbf{SFFNet-B (Ours)} & \textbf{35.9} & \textbf{58.2} & \textbf{37.3} & \textbf{28.6} & \textbf{46.6} & \textbf{46.2} & \textbf{13.6} & \textbf{39.0} & \textbf{51.0} & \textbf{44.1} & \textbf{61.2} & \textbf{65.3} & 19.7 & 105.6 \\
\midrule
\midrule
YOLOv8-L$^{\dagger}$ \cite{ref9} & 34.6 & 55.7 & 35.8 & 26.5 & 46.0 & 45.7 & 13.3 & 38.2 & 49.3 & 41.5 & \textbf{61.4} & 63.7 & 43.7 & 165.2 \\
YOLOv10-L$^{\dagger}$ \cite{ref16} & 34.4 & 56.0 & 35.9 & 26.8 & 45.1 & 42.2 & 13.0 & 37.6 & 49.3 & 42.5 & 59.7 & 61.8 & 24.4 & 120.3 \\
YOLO11-L$^{\dagger}$ \cite{ref17} & 34.5 & 55.6 & 35.7 & 26.6 & 45.5 & 45.8 & 13.2 & 38.0 & 49.2 & 42.0 & 60.5 & 60.9 & 25.3 & 86.9 \\
\rowcolor{gray!15}  
\textbf{SFFNet-L (Ours)} & \textbf{36.1} & \textbf{58.5} & \textbf{37.8} & \textbf{28.9} & \textbf{46.6} & \textbf{48.7} & \textbf{13.4} & \textbf{38.8} & \textbf{50.7} & \textbf{44.0} & 60.4 & \textbf{68.9} & 24.6 & 131.0 \\
\midrule
\midrule
YOLOv8-X$^{\dagger}$ \cite{ref9} & 35.6 & 57.3 & 37.2 & 27.2 & \textbf{47.6} & 44.5 & 13.5 & 38.9 & 50.1 & 42.4 & 62.0 & 65.5 & 68.2 & 257.8 \\
YOLOv10-X$^{\dagger}$ \cite{ref16} & 35.1 & 57.1 & 36.5 & 27.5 & 45.9 & 43.4 & 12.9 & 38.1 & 49.9 & 43.0 & 60.6 & 60.0 & 29.5 & 160.4 \\
YOLO11-X$^{\dagger}$ \cite{ref17} & 35.7 & 57.3 & 37.2 & 28.0 & 47.4 & 44.9 & 13.3 & 38.7 & 50.2 & 42.8 & \textbf{62.1} & 62.8 & 56.9 & 194.9 \\
\rowcolor{gray!15}  
\textbf{SFFNet-X (Ours)} & \textbf{36.8} & \textbf{59.3} & \textbf{38.5} & \textbf{29.6} & 47.4 & \textbf{45.4} & \textbf{13.6} & \textbf{39.5} & \textbf{51.4} & \textbf{44.8} & 61.1 & \textbf{65.6} & 38.5 & 203.7 \\
\bottomrule
\end{tabular}
}  
\end{table*}

\section{Experiments}
\label{sec:4}

\subsection{Datasets}
\subsubsection{VisDrone}
The VisDrone dataset \cite{ref12} is a large-scale benchmark for object detection and other computer vision tasks in drone images. Due to factors such as occlusions, small targets, uneven distribution, and significant scale changes in the data set, the object detection task is quite challenging. There are a total of 10,209 high-resolution images (2000 × 1500 pixels) in the dataset, including a training set (6,471 images), a validation set (548 images), and a test set (3,190 images). Among them, the test set is only available in specific competitions. In order to ensure the fairness and scientific nature of the experiment, the training set in this data set is used for model training, and the validation set is used to evaluate the model performance.

\subsubsection{UAVDT}
The UAVDT dataset \cite{ref11} is an important large-scale benchmark dataset widely used for UAV image object detection tasks. The dataset includes 50 video sequences captured from different locations in urban areas, covering various scene changes such as weather, perspective, and lighting, providing a comprehensive display of the complex and diverse characteristics of real-world scenes. The dataset contains three categories: car, truck, and bus.


\subsection{Implement Details}
Our models are trained using a single NVIDIA A100 GPU under the PyTorch framework. Unlike other models that rely on additional pre-training to gain performance advantages, all our models are trained from scratch without using pre-trained weights. The training phase is preset to 1,000 epochs, with an early stopping mechanism enabled (the model training automatically stops after it converges), and includes a warm-up phase of 3 epochs. The model uses the SGD optimizer, with SGD momentum and weight decay set to 0.937 and \(5\times10^{-4}\), respectively. The initial learning rate is \(1\times10^{-2}\) and decreases linearly to \(1\times10^{-4}\). For data augmentation, refer to \cite{ref9}, \cite{ref10}, and use random scaling, translation, and Mosaic \cite{ref7}, \cite{ref8}. During the training phase, the input size of each dataset sent to the model is fixed at 640 × 640, and the batch size is set to 8. During the testing phase, no multi-scale inference techniques are used for each dataset. The input size of the model is fixed at 1024 × 1024 for the VisDrone dataset and 640 × 640 for the UAVDT dataset.

\begin{table*}[!t]
\centering
\caption{\revised{Comparison of Other State-of-the-Art Models with Our Models on the Validation Set of the VisDrone Dataset. The \textcolor{red!40}{Red} Font Represents the Best Result Achieved for This Indicator, the \textcolor{green!55}{Green} Font Represents the Second-Best Result, and the \textcolor{gray!70}{Gray} Font Represents the Third-Best Result. “-” Indicates That the Corresponding Value Is Not Reported in the Relevant Paper. $^{\dagger}$ Denotes a Reimplementation of the Results under the Same Experimental Setting}
\label{tab:table2}} 
\renewcommand{\arraystretch}{1}  
\resizebox{0.8\textwidth}{!}{  
\begin{tabular}{l|c|c c c c c c|c c} 
\toprule
Model & Venue & \(\mathrm{AP}\) & \(\mathrm{AP}_{\mathrm{50}}\) & \(\mathrm{AP}_{\mathrm{75}}\) & \(\mathrm{AP}_{\mathrm{s}}\) 
& \(\mathrm{AP}_{\mathrm{m}}\) & \(\mathrm{AP}_{\mathrm{l}}\) & Param(M) & FLOPs(G) \\
\midrule
Faster-RCNN$^{\dagger}$ \cite{ref14} & NeurIPS 2015 & 24.7 & 40.5 & 26.2 & 15.7 & 36.9 & 37.9 & 41.4 & 208.0 \\
ClusDet \cite{ref39} & ICCV 2019 & 28.4 & 53.2 & 26.4 & 19.1 & 40.8 & \cellcolor{gray!20}54.4 & - & - \\
DMNet \cite{ref40} & CVPRW 2020 & 29.4 & 49.3 & 30.6 & 21.6 & 41.0 & \cellcolor{green!10}56.9 & - & - \\
GFL$^{\dagger}$ \cite{ref66} & NeurIPS 2020 & 26.4 & 42.7 & 27.7 & 17.4 & 38.3 & 45.6 & 52.3 & 283.0 \\
AMRNet \cite{ref67} & ArXiv 2020 & 32.1 & - & - & 23.2 & 43.9 & \cellcolor{red!15}60.5 & - & - \\
TOOD$^{\dagger}$ \cite{ref96} & ICCV 2021 & 27.5 & 44.0 & 28.8 & 18.6 & 39.1 & 43.2 & 32.0 & 199.0 \\
CDMNet \cite{ref68} & ICCVW 2021 & 30.7 & 51.3 & 32.0 & 22.2 & 42.4 & 44.7 & - & - \\
GLSAN \cite{ref69} & TIP 2021 & 32.5 & 55.8 & 30.0 & - & - & - & - & - \\ 
YOLOXs$^{\dagger}$ \cite{ref70} & ArXiv 2021 & 25.1 & 44.5 & 24.7 & 18.4 & 33.5 & 33 & 9.0 & 26.8 \\ 
QueryDet \cite{ref71} & CVPR 2022 & 28.3 & 48.1 & 28.8 & - & - & - & 40.0 & 499.0 \\ 
DetectoRS \cite{ref72} & ISPRS 2022 & 27.4 & 46.2 & 27.7 & 24.3 & 37.7 & - & - & - \\ 
SDP \cite{ref73} & TGRS 2023 & 30.2 & 52.5 & 30.6 & 22.6 & 39.6 & 39.8 & 97.0 & - \\ 
DDQ$^{\dagger}$ \cite{ref98} & CVPR 2023 & 34.4 & 56.8 & 35.1 & 26.0 & 45.4 & 52.0 & 48.3 & 274.0 \\ 
CZDet \cite{ref75} & CVPRW 2023 & 33.2 & 58.3 & 33.2 & 26.1 & 42.6 & 43.4 & - & - \\ 
EMA \cite{ref78} & ICASSP 2023 & 30.4 & 49.7 & - & - & - & - & 91.2 & 315.0 \\ 
PRDET \cite{ref79} & TCSVT 2023 & 32.0 & 53.9 & 33.2 & 25.6 & 40.8 & 52.9 & - & - \\ 
DINO$^{\dagger}$ \cite{ref97} & ICLR 2023 & 34.7 & 58.0 & 34.9 & 26.5 & 45.3 & 51.6 & 47.0 & 279.0 \\
EF-DETR \cite{ref82} & TII 2024 & 13.2 & 24.2 & 12.3 & 8.1 & 18.4 & 35.7 & - & - \\ 
BRSTD$^{\dagger}$ \cite{ref83} & TGRS 2024 & 28.3 & 48.1 & 28.4 & 23.2 & 34.8 & 24.6 & 1.8 & 64.0 \\ 
BRSTD-l$^{\dagger}$ \cite{ref83} & TGRS 2024 & 31.4 & 52.8 & 32.3 & 26.5 & 38.5 & 27.5 & 6.3 & 219.6 \\ 
NE-CDMNet \cite{ref58} & TGRS 2024 & \cellcolor{gray!20}35.9 & \cellcolor{red!15}59.4 & 35.9 & 27.3 & \cellcolor{green!10}46.6 & 53.1 & 76.0 & 86.0 \\ 
RT-DETR$^{\dagger}$ \cite{ref81} & CVPR 2024 & 25.2 & 42.9 & 25.0 & 17.1 & 34.8 & 42.1 & 42.0 & 136.0 \\ 
DDOD$^{\dagger}$ \cite{ref99} & TMM 2024 & 23.0 & 38.3 & 23.6 & 14.3 & 34.2 & 39.5 & 32.2 & 180.0 \\ 
SDPDet \cite{ref95} & TMM 2024 & 33.7 & 56.6 & 34.3 & 26.7 & 42.9 & 45.7 & - & 139.7 \\
BAFNet \cite{ref84} & TGRS 2025 & 30.8 & 52.6 & 30.9 & 22.4 & 41.0 & 43.2 & - & - \\
\midrule
\textbf{SFFNet-N (Ours)} & - & 26.7 & 45.4 & 26.8 & 19.8 & 35.3 & 35.0 & 1.7 & 7.2 \\
\textbf{SFFNet-S (Ours)} & - & 31.6 & 52.8 & 32.3 & 25.3 & 40.8 & 35.7 & 6.3 & 24.0 \\
\textbf{SFFNet-M (Ours)} & - & 35.3 & 57.5 & 36.6 & 28.0 & \cellcolor{gray!20}46.0 & 42.9 & 14.2 & 63.2 \\
\textbf{SFFNet-B (Ours)} & - & \cellcolor{gray!15}35.9 & 58.2 & \cellcolor{gray!20}37.3 & \cellcolor{gray!20}28.6 & \cellcolor{green!10}46.6 & 46.2 & 19.7 & 105.6 \\
\textbf{SFFNet-L (Ours)} & - & \cellcolor{green!10}36.1 & \cellcolor{gray!20}58.5 & \cellcolor{green!10}37.8 & \cellcolor{green!10}28.9 & \cellcolor{green!10}46.6 & 48.7 & 24.6 & 131.0 \\
\textbf{SFFNet-X (Ours)} & - & \cellcolor{red!15}36.8 & \cellcolor{green!10}59.3 & \cellcolor{red!15}38.5 & \cellcolor{red!15}29.6 & \cellcolor{red!15}47.4 & 45.4 & 38.5 & 203.7 \\
\bottomrule
\end{tabular} }
\end{table*}

\subsection{Evaluation Metrics}
The evaluation protocol of the MS-COCO \cite{ref13} dataset is adopted in this article. First, we adopt \({AP}\) as the primary metric, used to measure the overall performance of the detection results. The specific definition is as follows
\begin{equation}
    AP=\frac{1}{n}\sum_{k=1}^nJ(P,R)_k,
\end{equation}
where \({n}\) represents the total number of object categories, and \(J(P,R)_k\) denotes the AP function for each category. \({P}\) and \({R}\) represent precision and recall, respectively. The formulas for precision and recall are as follows
\begin{equation}
    P=\frac{TP}{TP+FP},
\end{equation}
\begin{equation}
    R=\frac{TP}{TP+FN},
\end{equation}
where \({TP}\) is the number of true positives, \({FP}\) is the number of false positives, and \({FN}\) is the number of false negatives.

Specifically, \(AP_{50}\) represents the \(AP\) value when the IoU threshold is 0.50, and \(AP_{75}\) represents the \(AP\) value when the IoU threshold is 0.75, corresponding to more lenient and stricter detection standards, respectively. Additionally, we also evaluate the model's detection performance at different object scales. Specifically, \(AP_{s}\), \(AP_{m}\), and \(AP_{l}\) represent the average precision for small, medium, and large objects, respectively. To further quantify the coverage of the model in object detection, this article uses \(AR\) as a supplementary metric.

\subsection{Comparison with State-of-the-Art Methods}
\revised{In this section, we primarily compared our model with the YOLO series detectors under the same experimental conditions to correctly assess the relative advantages of our model. Additionally, we also selected several detectors that have demonstrated outstanding performance in both classic object detection and small object detection to conduct benchmark testing, further highlighting the absolute competitiveness of our model, particularly its excellent performance in complex scenes and small object detection.}

\begin{table}[!t]
\centering
\caption{\revised{Comparison of Other State-of-the-Art Models with Our Models on the UAVDT Dataset. The Best Result Is Marked in Bold.  “-” Indicates That the Corresponding Value Is Not Reported in the Relevant Paper. $^{\dagger}$ Denotes a Reimplementation of the Results under the Same Experimental Setting}
\label{tab:table3}} 
\renewcommand{\arraystretch}{1}  
\resizebox{0.45\textwidth}{!}{  
\begin{tabular}{c||c c c | c c c} 
\toprule
Model & \(\mathrm{AP}\) & \(\mathrm{AP}_{\mathrm{50}}\) & \(\mathrm{AP}_{\mathrm{75}}\) & \(\mathrm{AP}_{\mathrm{s}}\) 
& \(\mathrm{AP}_{\mathrm{m}}\) & \(\mathrm{AP}_{\mathrm{l}}\) \\
\midrule
ClusDet \cite{ref39} & 13.7 & 26.5 & 12.5 & 9.1 & 25.1 & 31.2 \\
CenterNet \cite{ref85}& 13.2 & 26.7 & 11.8 & 7.8 & 26.6 & 13.9 \\
DMNet \cite{ref40} & 14.7 & 24.6 & 16.3 & 9.3 & 26.2 & \textbf{35.2} \\
GFL$^{\dagger}$ \cite{ref66} & 17.0 & 30.3 & 18.1 & 11.2 & 28.5 & 30.9 \\
AMRNet \cite{ref67} & 18.2 & 30.4 & 19.8 & 10.3 & 31.3 & 33.5 \\
GLSAN \cite{ref69} & 19.0 & 30.5 & 21.7 & - & - & - \\
CEASC \cite{ref74} & 17.1 & 30.9 & 17.8 & - & - & - \\
PRDET \cite{ref79} & 19.1 & 33.8 & 19.8 & - & - & - \\
SCLNet \cite{ref59} & 20.0 & 33.1 & \textbf{22.3} & - & - & - \\
YOLOv8-X$^{\dagger}$ \cite{ref9} & 18.1 & 31.1 & 18.8 & 13.0 & 28.4 & 32.4 \\
YOLOv10-X$^{\dagger}$ \cite{ref16} & 19.5 & 32.3 & 21.0 & 13.4 & 28.9 & 31.5 \\
YOLO11-X$^{\dagger}$ \cite{ref17} & 19.0 & 33.0 & 19.6 & 13.0 & 30.7 & 32.2 \\
\midrule
\textbf{SFFNet-X (Ours)} & \textbf{20.6} & \textbf{34.4} & 21.9 & \textbf{14.3} & \textbf{31.5} & 33.3 \\
\bottomrule
\end{tabular} }
\end{table}

\subsubsection{Experiments on VisDrone}
\revised{As shown in Table~\ref{tab:table1}, SFFNet demonstrates strong competitiveness across various model scales and multiple object detection metrics. First, we compared SFFNet with the baseline model (YOLOv10 \cite{ref16}). Among the six variants (N/S/M/B/L/X), SFFNet improved \(AP\) by 3.8\%, 4.4\%, 3.4\%, 2.1\%, 1.7\%, and 1.7\%, respectively, and improved \(AP_{s}\) by 2.7\%, 4.5\%, 3.3\%, 2.3\%, 2.1\%, and 2.1\%, respectively. Compared to other YOLO models, our model also demonstrates a superior balance between accuracy and computational cost. Specifically, for lightweight and small models, SFFNet-N/S outperforms YOLOv8-N/S by 3\%/3.8\% in \(AP\), and by 3\%/4.7\% in \(AP_{s}\), while reducing the number of parameters by 46\%/43\%, respectively. SFFNet-N/S also outperforms YOLO11-N/S by 2.8\%/3.8\% in \(AP\), and by 2.9\%/4.8\% in \(AP_{s}\), with the number of parameters decreasing by 34\% and 32\%, respectively. For medium-sized models, SFFNet-M offers higher performance while reducing the number of parameters by 45\% and 29\% compared to YOLOv8-M and YOLO11-M, respectively. For large models, SFFNet-L reduces the number of parameters by 43\% and 2\% compared to YOLOv8-L and YOLO11-L, respectively, while achieving significant improvements in \(AP\) (1.5\% and 1.6\%, respectively). Specifically, SFFNet-X further compresses the number of parameters and reduces the computational cost by 20\% compared to YOLOv8-X, and is almost on par with YOLO11-X. In this case, \(AP\) is further improved by 1.2\% and 1.1\%, respectively.}

As shown in Table~\ref{tab:table2}, the quantitative detection performance results of several state-of-the-art methods are presented, compared to our proposed model on the VisDrone validation set. The results show that our model outperforms existing methods on most evaluation metrics. It is particularly worth noting that traditional general object detection frameworks, like Faster-RCNN \cite{ref14}, are limited by the unique characteristics of aerial images, including the dense distribution of small objects and scale imbalance, with an \(AP\) of only 24.8\%. In contrast, the density map-based clipping method (such as NE-CDMNet \cite{ref58}) shows a significant improvement in detection performance, with an \(AP\) of 35.9\%. In the top three detection results, our series of models occupy almost all of the positions. It is somewhat regrettable that our model did not make it into the top three in terms of large object detection metrics. Nevertheless, this is acceptable, as the majority of objects in aerial images tend to be small or medium-sized. Hence, enhancing the detection capabilities for these object sizes will significantly improve overall performance, aligning more closely with the practical application needs of real-world scenarios.

It is worth noting that transformer-based DETR \cite{ref15} variants, despite their strong performance on COCO and other general benchmarks, exhibit notably inferior results on VisDrone. For instance, our reimplemented RT-DETR \cite{ref81} achieves only 25.2\% AP, which is lower than its CNN-based counterparts. This phenomenon is primarily attributed to the mismatch between RT-DETR's architectural assumptions and UAV-specific scene characteristics. First, UAV images are dominated by extremely small objects requiring fine-grained feature representation; however, most DETR variants typically initiate feature interaction from deeper layers, lacking explicit preservation mechanisms for high-resolution shallow features that are critical for recovering the shape and texture of tiny objects. Second, although the sparse query mechanism is efficient for general detection, it faces physical limitations in spatial coverage when handling the high-density crowds typical of aerial scenes, leading to potential query competition. Finally, regarding the training paradigm, while large-scale pretraining (e.g., on COCO) usually benefits transformers, the semantic priors learned from general objects cannot fully compensate for the loss of high-frequency spatial details required for UAV small object detection. Consequently, the global self-attention mechanism, without targeted inductive biases for local edge enhancement, struggles to capture the sparse pixel information of small targets as effectively as our proposed approach. Our SFFNet achieves 36.8\% AP through MDDC's dual-domain edge enhancement for improved small object clarity and SFPN's high-resolution feature preservation. These results indicate that CNN architectures with explicit multi-scale fusion and small object enhancement remain more suitable for resource-constrained UAV detection tasks, although we acknowledge that ongoing DETR research may narrow this gap in the future.

\begin{table}[!t]
\centering
\caption{\revised{Ablation Accuracy Comparison between the Baseline Model and the Models of Different Components in Our Method on the VisDrone Dataset Validation Set, Where w/o DEIE Represents the MDDC Module without the DEIE Component}}
\label{tab:table4}
\renewcommand{\arraystretch}{1.2}
\resizebox{0.49\textwidth}{!}{
\begin{tabular}{c||cc|cc|cccccc|cc}
\toprule
& \multicolumn{2}{c|}{MDDC} & \multicolumn{2}{c|}{SFPN}
& \multirow{2}{*}{AP}
& \multirow{2}{*}{AP$_{50}$}
& \multirow{2}{*}{AP$_{75}$}
& \multirow{2}{*}{AP$_{\mathrm{s}}$}
& \multirow{2}{*}{AP$_{\mathrm{m}}$}
& \multirow{2}{*}{AP$_{\mathrm{l}}$}
& \multirow{2}{*}{Param(M)}
& \multirow{2}{*}{FLOPs(G)} \\
\cmidrule(lr){2-3}\cmidrule(lr){4-5}
& DEIE & w/o DEIE & LDConv & WPM & & & & & & & & \\
\midrule
I   &  &            &         &         & 27.2 & 46.2 & 27.5 & 20.8 & 35.6 & 31.9 & 7.2 & 21.6 \\
II  &  & \checkmark &         &         & 27.8 & 47.1 & 28.2 & 20.9 & 37.1 & 32.1 & 7.4 & 20.0 \\
III & \checkmark & \checkmark &       &     & 28.1 & 47.6 & 28.4 & 21.3 & 37.2 & 32.5 & 7.4 & 20.1 \\
IV  &  &            &         & \checkmark  & 29.4 & 49.6 & 29.9 & 22.9 & 38.4 & 31.3 & 5.8 & 22.1 \\
V   &  &            & \checkmark & \checkmark & 31.3 & 52.0 & 31.8 & 24.6 & 40.3 & 37.1 & 6.1 & 25.4 \\
VI  &  & \checkmark & \checkmark & \checkmark & 31.3 & 52.3 & 31.9 & 24.8 & 40.7 & 34.9 & 6.3 & 23.7 \\
VII & \checkmark & \checkmark &         & \checkmark & 30.3 & 50.8 & 30.7 & 24.0 & 39.5 & 33.8 & 6.0 & 20.6 \\
VIII& \checkmark & \checkmark & \checkmark &     & 28.9 & 48.9 & 29.3 & 22.6 & 37.8 & 31.6 & 5.9 & 18.5 \\
IX  & \checkmark & \checkmark & \checkmark & \checkmark & \textbf{31.6} & \textbf{52.8} & \textbf{32.3} & \textbf{25.3} & \textbf{40.8} & \textbf{35.7} & 6.3 & 24.0 \\
\bottomrule
\end{tabular}}
\end{table}

\subsubsection{Experiments on UAVDT}
The UAVDT dataset contains a large number of small objects and includes many images with low lighting and complex backgrounds, which more accurately reflect the network's performance in small object detection tasks. As shown in Table~\ref{tab:table3}, compared to the baseline YOLOv10-X, our method improves the \(AP_{50}\) by 2.1\%, surpassing many existing state-of-the-art methods. Compared to the latest method, SCLNet \cite{ref59}, our detection performance improves by 0.6\% and 1.3\% in the \(AP\) and \(AP_{50}\) metrics, respectively. These results demonstrate that our method performs excellently in aerial image object detection task.

\subsection{Ablation Experiments and Analysis}
\begin{figure}[!t]   
\captionsetup{skip=0pt}  
\centering
\includegraphics[width=\columnwidth]{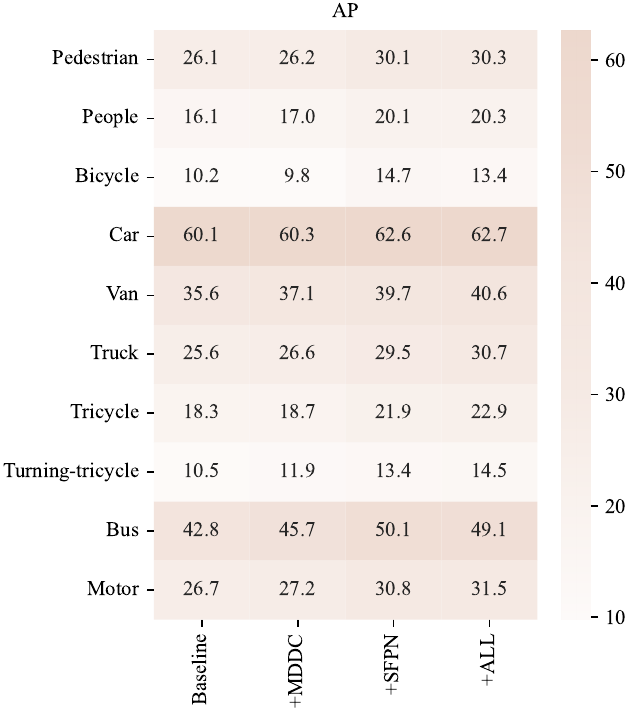}  
\caption{The qualitative results of the ablation experiment for all fine-grained categories on the validation set of the VisDrone dataset. Deeper colors indicate higher AP values and better detection performance. "+All" indicates the addition of all components.}
\label{fig_7}
\end{figure}

In this section, we present a series of comprehensive ablation experiments conducted on the VisDrone dataset. Unless otherwise indicated, the experiments were performed on the SFFNet-S model. \revised{The primary objective of these experiments is to evaluate the individual contributions of the proposed MDDC and SFPN components to the overall model performance, including fine-grained ablation of their internal sub-components and the synergistic enhancement achieved through their joint integration.} As shown in Table~\ref{tab:table4}, in these experiments, each component is compared with the baseline model, \revised{and Fig.~\ref{fig_7} further illustrates the ablation experiment performance across multiple fine-grained categories. In Table~\ref{tab:table5}, we also compared with several common and competitive feature pyramid networks to demonstrate the superiority of SFPN as the neck network.}

\subsubsection{Effectiveness of the MDDC}
Aerial images encompass objects of varying scales, with smaller objects often obscured by intricate backgrounds and subjected to high levels of noise. When directly applied to evaluate performance on the VisDrone dataset, the Baseline yields suboptimal results, particularly for smaller object scales. This can be attributed to the Baseline's difficulty in effectively distinguishing between the foreground and background, which hampers its ability to extract relevant features and accurately localize smaller objects. Fortunately, upon the introduction of the MDDC component, the model focuses on the multi-scale edge information of objects, effectively suppressing complex background noise. \revised{As shown in the third row of Table~\ref{tab:table4}, \(AP_{s}\), \(AP_{m}\), and \(AP_{l}\) increase by 0.5\%, 1.6\%, and 0.6\%, respectively, compared to the baseline. Additionally, \(AP\) rises from 27.2\% to 28.1\%. }

\revised{To further investigate the functional interplay of the dual-domain mechanism in the DEIE module, we conducted detailed ablation experiments (Table~\ref{tab:mddc_ablation}). The results show that using only the frequency-domain branch increases the \(AP\) from 27.2\% to 27.7\%, while using only the spatial-domain branch raises it to 27.5\%. When both domains are combined (without edge strength map guidance), the \(AP\) further improves to 27.9\%, and the complete configuration (with edge strength map guidance) reaches 28.1\%. This progressive improvement demonstrates the complementarity between the frequency and spatial domains, enabling the model to suppress large-scale background noise while maintaining precise spatial localization accuracy.} \revised{As evidenced by the comparison between the second and third rows, the incorporation of the DEIE component leads to further performance enhancement, primarily attributed to the synergistic interplay between frequency-domain and spatial-domain processing that amplifies fine-grained features and improves small-object detection clarity.} At the same time, the model’s computational cost is reduced by 1.5~G, while the number of parameters experiences only a slight increase of 0.2~M, thereby achieving an effective balance between detection accuracy and model efficiency. The strategic utilization of edge frequency-domain information is the key factor for the competitive results achieved by the MDDC component.

\begin{table}[!t]
\centering
\caption{\revised{Detailed ablation study of DEIE components within the MDDC module}}
\label{tab:mddc_ablation}
\renewcommand{\arraystretch}{1.3}
\resizebox{0.49\textwidth}{!}{
\begin{tabular}{l||ccc|cccc}
\toprule
\textbf{Configuration} 
& \textbf{Frequency} 
& \textbf{Spatial} 
& \textbf{Edge strength map} 
& \textbf{AP} 
& \textbf{AP$_{\mathrm{s}}$} 
& \textbf{AP$_{\mathrm{m}}$} 
& \textbf{AP$_{\mathrm{l}}$} \\
\midrule
Baseline 
& $\times$ & $\times$ & $\times$ 
& 27.2 & 20.8 & 35.6 & 31.9 \\
+ Frequency domain 
& \checkmark & $\times$ & $\times$ 
& 27.7 & 21.1 & 36.2 & 32.1 \\
+ Spatial domain 
& $\times$ & \checkmark & $\times$ 
& 27.5 & 21.0 & 36.0 & 31.8 \\
+ Dual domain 
& \checkmark & \checkmark & $\times$ 
& 27.9 & 21.2 & 36.8 & 32.3 \\
+ All 
& \checkmark & \checkmark & \checkmark 
& \textbf{28.1} & \textbf{21.3} & \textbf{37.2} & \textbf{32.5} \\
\bottomrule
\end{tabular}}
\end{table}

\begin{table}[!t]
\centering
\caption{\revised{Ablation Study of Different WPM Internal Convolution Kernel Configurations}}
\label{tab:wpm_ablation2}
\renewcommand{\arraystretch}{1.2}
\resizebox{0.49\textwidth}{!}{
\begin{tabular}{c||c||cccccc|cc}
\toprule
& \textbf{WPM} 
& \textbf{AP} 
& \textbf{AP$_{50}$} 
& \textbf{AP$_{75}$} 
& \textbf{AP$_{\mathrm{s}}$} 
& \textbf{AP$_{\mathrm{m}}$} 
& \textbf{AP$_{\mathrm{l}}$} 
& \textbf{Param(M)} 
& \textbf{FLOPs(G)} \\
\midrule
I & $1{\times}1,\;1{\times}3,\;3{\times}1,\;3{\times}3$ 
& 31.3 & 52.1 & 32.2 & 25.0 & 41.1 & 34.0 & 6.2 & 22.6 \\
II & $1{\times}1,\;1{\times}7,\;7{\times}1,\;7{\times}7$ 
& 31.1 & 51.8 & 32.1 & 24.8 & 40.7 & 34.9 & 6.2 & 22.6 \\
III & $1{\times}1,\;1{\times}15,\;15{\times}1,\;15{\times}15$
& 31.4 & 52.4 & 32.2 & 25.0 & 40.8 & 35.3 & 6.2 & 22.9 \\
IV & $1{\times}1,\;1{\times}31,\;31{\times}1,\;31{\times}31$
& 31.6 & 52.8 & 32.3 & 25.3 & 40.8 & 35.7 & 6.3 & 24.0 \\
V & $1{\times}1,\;1{\times}63,\;63{\times}1,\;63{\times}63$
& 31.7 & 53.0 & 32.4 & 25.4 & 41.0 & 35.8 & 6.6 & 27.6 \\
\bottomrule
\end{tabular}}
\end{table}

\revised{In Fig.~\ref{fig_7}, we further present the ablation comparison results for fine-grained categories after adding the MDDC component. Compared to the first column, the \(AP\) for several categories has improved. Nevertheless, the performance of the bicycle category has slightly declined due to the complexity of its hollow contours and intricate edges, which are often densely packed in urban environments. Capturing edge frequency domain features of these closely arranged structures is challenging. Moreover, bicycles are often associated with riders, causing confusion between their edge features and leading the model to focus more on the rider's features, reducing accuracy in representing the bicycle.}

\begin{table*}[!t]
\centering
\caption{Comparison of the Performance of SFPN and Other Feature Pyramids Networks on the VisDrone Validation Set
\label{tab:table5}} 
\renewcommand{\arraystretch}{1.25}  
\resizebox{0.8\textwidth}{!}{  
\begin{tabular}{c|c c c|c c c|c c c|c c c|c c} 
\toprule
Model & \(\mathrm{AP}\) & \(\mathrm{AP}_{\mathrm{50}}\) & \(\mathrm{AP}_{\mathrm{75}}\) & \(\mathrm{AP}_{\mathrm{s}}\) 
& \(\mathrm{AP}_{\mathrm{m}}\) & \(\mathrm{AP}_{\mathrm{l}}\) & \(\mathrm{AR}_{\mathrm{1}}\) & \(\mathrm{AR}_{\mathrm{10}}\) & \(\mathrm{AR}_{\mathrm{100}}\) & \(\mathrm{AR}_{\mathrm{s}}\) & \(\mathrm{AR}_{\mathrm{m}}\) & \(\mathrm{AR}_{\mathrm{l}}\) & Param(M) & FLOPs(G) \\
\midrule
PAFPN \cite{ref4} & 27.2 & 46.2 & 27.5 & 20.8 & 35.6 & 31.9 & 10.7 & 31.9 & 43.2 & 36.7 & 52.5 & 46.5 & 7.2 & 21.6 \\
BIFPN \cite{ref1} & 28.6 & 48.1 & 29.1 & 22.2 & 37.4 & 31.7 & 11.1 & 33.1 & 44.2 & 37.6 & 54.3 & 44.7 & 5.2 & 18.7 \\
AFPN \cite{ref86} & 27.7 & 46.1 & 28.4 & 20.6 & 37.3 & 36.4 & 11.0 & 32.3 & 42.6 & 35.3 & 54.0 & 51.1 & 7.6 & 24.2 \\
RepGFPN \cite{ref87} & 27.8 & 47.1 & 28.3 & 21.4 & 36.2 & 34.9 & 10.9 & 32.4 & 43.7 & 37.2 & 53.5 & 47.7 & 9.6 & 23.2 \\
MAFPN \cite{ref88} & 29.1 & 48.9 & 29.5 & 22.2 & 38.4 & 35.2 & 11.2 & 33.5 & 44.7 & 37.5 & 55.1 & 47.1 & 10.3 & 26.8 \\
\midrule
\textbf{SFPN (Ours)}& \textbf{31.3} & \textbf{52.0} & \textbf{31.8} & \textbf{24.6} & \textbf{40.3} & \textbf{37.1} & \textbf{12.0} & \textbf{35.1} & \textbf{46.7} & \textbf{40.3} & \textbf{56.4} & \textbf{53.7} & 6.1 & 25.4 \\
\bottomrule
\end{tabular} }
\end{table*}

\begin{figure*}[!t]   
\captionsetup{skip=0pt}  
\centering
\includegraphics[width=0.8\textwidth]{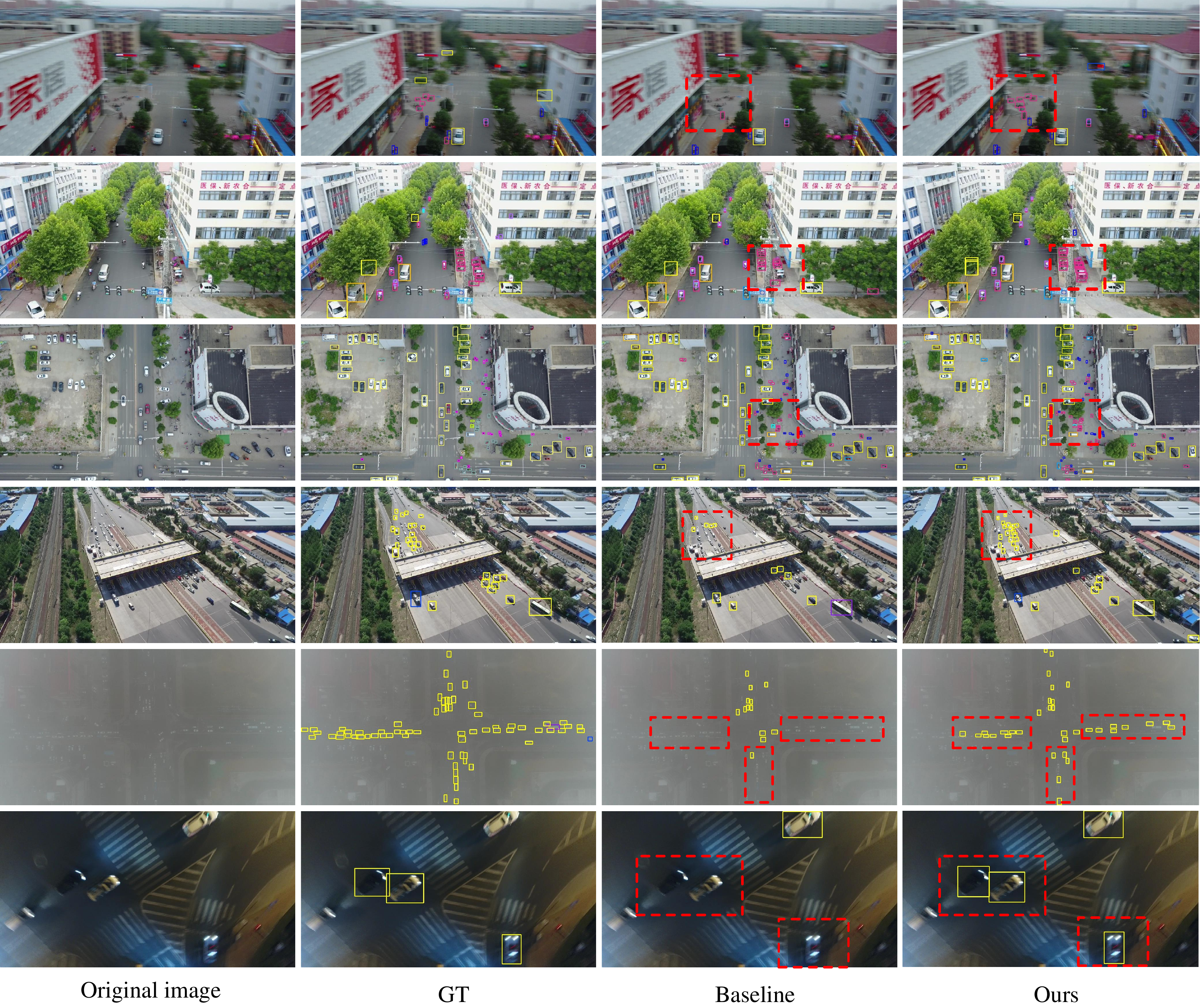}  
\caption{The comparison of detection visualization results between the baseline model and SFFNet. The first three rows are from the VisDrone dataset, and the last three are from the UAVDT dataset. Our model performs better in handling dense areas from various perspectives (rows 2–4), noise occlusion resulting from specific weather conditions (row 5), and blur caused by camera rotation (rows 1 and 6).}
\label{fig_8}
\end{figure*}

\subsubsection{Effectiveness of the SFPN}
After the original features are extracted by the backbone network, they are fed into the neck network, where the SFPN enables the fusion and enhancement of features, specifically optimized for the detection of small objects in drone images. \revised{As shown in the fifth row of Table~\ref{tab:table4}, compared to the baseline, \(AP\) increases from 27.2\% to 31.3\%, and \(AP_{s}\), \(AP_{m}\), and \(AP_{l}\) increase by 3.8\%, 4.7\%, and 5.2\%, respectively. More specifically, we conducted fine-grained ablation studies targeting two pivotal sub-components within SFPN, LDConv and WPM. The row-wise comparisons in the table provide further evidence of their respective contributions to the overall performance.} These experiments fully demonstrate that SFPN, in the fusion mechanism, preserves feature information from all layers, avoiding the suppression of low-level features and ensuring that spatial information and details are fully transmitted. Furthermore, it can be seen that SFPN also further enhances the ability to extract low-level features, making the capture of fine-grained features for small objects more accurate. \revised{Detailed reports for specific categories are shown in Fig.~\ref{fig_7}. The third column shows the results of adding only the SFPN component to the baseline, with strong performance in the \(AP\) of all ten categories. Notably, the detection of the bus category is especially outstanding, as buses typically exhibit significant scale variation and complex backgrounds in images. SFPN effectively enhances detection ability through multi-scale feature fusion, resulting in the largest AP improvement for buses.}

\begin{figure*}[htbp]   
\captionsetup{skip=0pt}  
\centering
\includegraphics[width=0.85\textwidth]{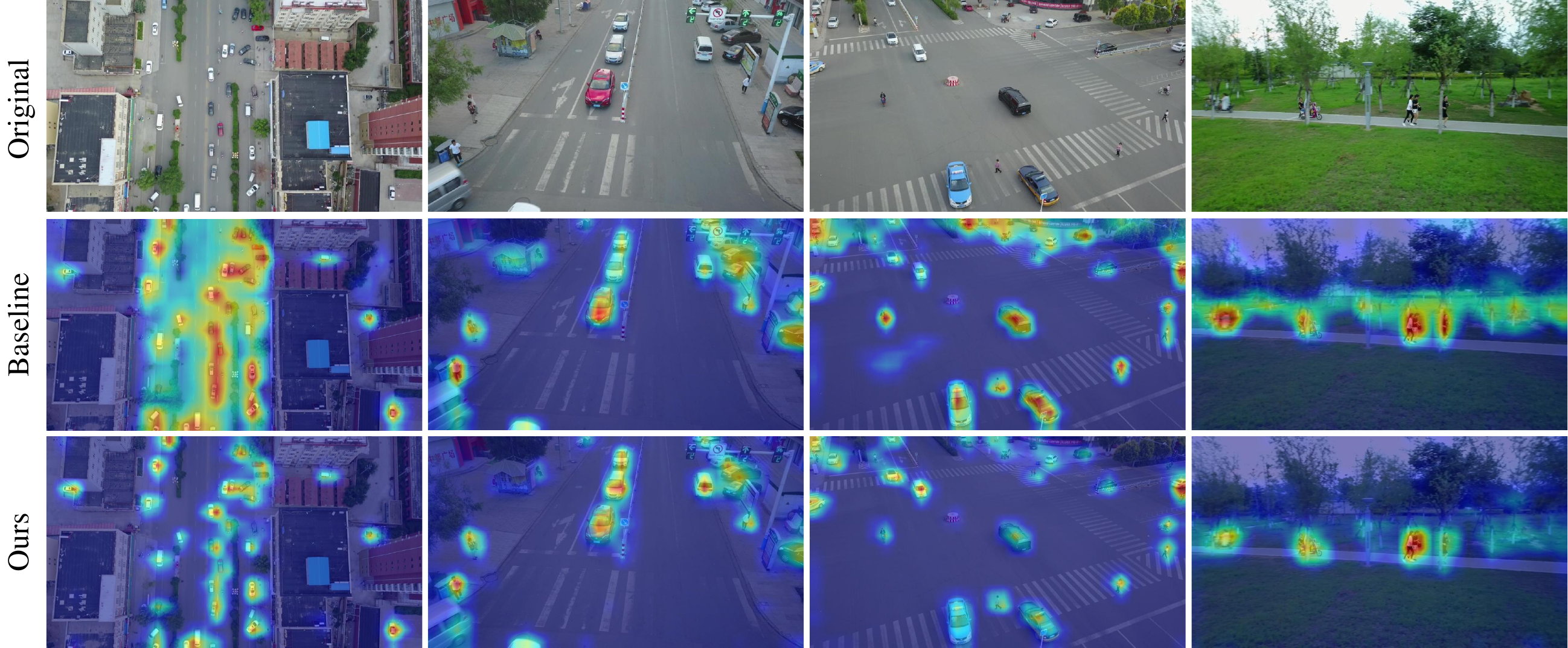}  
\caption{The comparison of heatmap visualization results between the baseline model and SFFNet on the VisDrone dataset. Our model not only suppresses irrelevant background noise but also focuses more on detecting the target objects.}
\label{fig_9}
\end{figure*}

\subsubsection{Kernel Size Selection for WPM}
\revised{To verify the rationality of selecting the \(31 \times 31\) kernel, we evaluated five kernel-size configurations, as shown in Table~\ref{tab:wpm_ablation2}. The suboptimal performance of smaller kernels (\(3\times3\) and \(7\times7\)) can be attributed to their limited receptive fields on the P3 layer (\(80\times80\) feature map), which are insufficient to capture adequate contextual information for small objects. For instance, for a \(10\times10\) pixel vehicle, a \(3\times3\) kernel can only capture a local fragment, making it difficult to establish correlations between the target and its surroundings. Although a \(7\times7\) kernel is slightly larger, it falls into an “intermediate” range that is neither sufficiently local nor adequately global, leading to further performance degradation.}

\revised{When the kernel size increases to \(15\times15\), performance begins to recover because the receptive field can now cover medium-scale targets along with their nearby context. The \(31\times31\) kernel further improves the overall \(AP\) to \(31.6\%\), offering a sufficiently wide receptive field for densely distributed small-object scenes and enabling the perception of target-group distribution patterns. Enlarging the kernel to \(63\times63\) yields only a marginal \(0.1\%\) increase in \(AP\) (to \(31.7\%\)) while FLOPs jump from \(24.0\,\mathrm{G}\) to \(27.6\,\mathrm{G}\) (+15\%). Although \(63\times63\) shows a slight advantage in small-object detection, on the \(80\times80\) feature map it suffers from severe boundary overflow, causing excessive padding and computational waste. Moreover, an overly large receptive field leads to feature over-smoothing, weakening local details. Considering both performance and efficiency, the \(31\times31\) kernel represents the optimal choice, maintaining an excellent balance for small and medium targets (\(AP_{s}\)=25.3\%, \(AP_{m}\)=40.8\%).}



\subsection{SFPN vs. Other FPNs}
In Table~\ref{tab:table5}, we integrate different FPNs into the baseline model YOLOv10-S and present a performance comparison in the UAV image object detection task to further highlight the competitive performance of SFPN. Specifically, SFPN achieves higher values than other FPNs in terms of \(AP\), \(AP_{50}\), and \(AP_{75}\) metrics. Notably, compared to the baseline model using PAFPN \cite{ref4}, the increase in \(AP_{50}\) reaches 6.4\%, while the number of parameters decreases by 1.1M, and the computational cost only increases by 3.8G. The comparison results show that SFPN has excellent performance capabilities in small objects and complex scenarios, which fully verifies the rationality of our SFPN design ideas.

\subsection{Visualization of Detection Results}
To more clearly demonstrate the superiority of the proposed method, Fig.~\ref{fig_8} presents visualizations of detection results from both the baseline model (YOLOv10-X) and the SFFNet-X model on the VisDrone and UAVDT datasets. The annotated bounding boxes in the original images are used as the standard reference. Specifically, from the drone's perspective, small objects at a long distance appear densely arranged, resulting in more complex features in these areas. In Fig.~\ref{fig_8}, the camera's motion in the first and last rows causes significant changes in perspective, leading to image blur and insufficient feature representation of the objects under the regular viewpoint. In rows two to four, aerial images often contain dense areas and small objects. The fifth row shows the impact of heavy fog and high flight angles, causing the source image to present a broader background noise. Based on the aforementioned real-world issues, the comparison results from each row in the figures demonstrate that our model demonstrates significant advantages in complex scenarios, especially in cases with small or dense objects, object occlusion, or image blur. It effectively reduces background noise and focuses on dense target areas. At the same time, this strongly validates the crucial role of the MDDC and SFPN components in enhancing the model's performance.

\subsection{Visualization of Heatmaps}
Fig.~\ref{fig_9} presents heatmaps generated by SFFNet-X on the VisDrone dataset, visually illustrating the attention that the proposed method pays to object areas and key patterns. In comparison with the baseline model (YOLOv10-X), our SFFNet-X model demonstrates a significant improvement in suppressing irrelevant background noise. For instance, the trees on both sides of the road and the empty regions in the center of the road are nearly entirely excluded. The strategic application of dual-domain edge technology in MDDC plays a pivotal role in enabling the model to mitigate the influence of background noise, thereby enhancing its capability to detect objects with heightened focus and precision.

\section{Conclusion}
\label{sec:5}
This article introduces SFFNet, a novel end-to-end model specifically tailored for object detection in UAV aerial imagery. We first proposed the MDDC module to address challenges related to limited target feature representation and background noise interference. Additionally, we explored the potential of feature pyramids in the neck for processing object geometric and long-range contextual information, achieving synergistic multi-scale object modeling through SFPN. Quantitative results on the VisDrone and UAVDT datasets demonstrate that SFFNet excels in UAV aerial image object detection tasks, achieving competitive detection accuracy. Qualitative analyses further reveal that SFFNet outperforms baseline methods by effectively suppressing low-frequency noise interference in aerial scenes and exhibiting enhanced robustness in multi-scale object detection. However, there is still room for improvement in detecting large objects. In the future, we will explore adaptive anchor box generation mechanisms that dynamically adjust anchor box scales based on object sizes, in order to enhance detection accuracy for large objects while maintaining precise detection of small ones.

\bibliographystyle{IEEEtran}
\bibliography{REF}

\vfill

\end{document}